\documentclass[journal]{IEEEtran}

\usepackage[latin1]{inputenc}
\usepackage{url}                                                   % paper
\usepackage{times}
\usepackage{epsfig}
\usepackage{graphicx}
\usepackage{amsmath}
\usepackage{amssymb}
\usepackage{multirow}
\usepackage{multicol}
\usepackage{array}
\usepackage{rotating}
\usepackage{placeins}
\newcommand{\ie}{i.e.,} %del latin.. this is.
\newcommand{\eg}{e.g.,} %del latin... for example
\newcommand{\etal}{\emph{et al. }} %del latin... for example

\newcommand{\tab}[1]{Table~\ref{#1}}
\newcommand{\fig}[1]{Fig. \ref{#1}}

\newcommand{\subfig}[2]{Fig.~\ref{#1}#2}

\newcommand{\sect}[1]{Sect. \ref{#1}}
\newcommand{\eq}[1]{Eq. (\ref{#1})}

\newcommand{\argmax}{\operatornamewithlimits{argmax}}
\newcommand{\argmin}[1]{\underset{#1}{\operatorname{argmin}}\;}

\DeclareGraphicsExtensions{.eps,.ps}
\graphicspath{{./figsTesis/}}

\begin{document}
\title{Road Detection via On--line Label Transfer}
\author{Jos\'e M. \'Alvarez, Ferran Diego, Joan Serrat and Antonio M. L\'opez
\thanks{The authors are with Computer Vision Center \& Computer Science Dept., Edifici O, Universitat Autònoma de Barcelona, 08193 Cerdanyola del Vallés, Spain. {\tt\small jalvarez@cvc.uab.es}}}
\author{Jose~M.~Alvarez,
	Ferran~Diego,
        Joan~Serrat,
        Antonio~M.~L\'opez% <-this % stops a space
\IEEEcompsocitemizethanks{\IEEEcompsocthanksitem This work is supported by Spanish MINECO projects TRA2011-29454-C03-01,
TIN2011-25606, TIN2011-29494-C03-02, and the Research Program Consolider Ingenio 2010: MIPRCV (CSD200700018) and the Catalan Generalitat project CTP-2008ITT00001..
\IEEEcompsocthanksitem The authors are with Computer Vision Center \& Computer Science Dept., Edifici O, Universitat Autònoma de Barcelona, 08193 Cerdanyola del Vallés, Spain. {\tt\footnotesize jalvarez@cvc.uab.es}}% <-this % stops a space
}

% The paper headers
\markboth{Journal of \LaTeX\ Class Files,~Vol.~6, No.~1, January~2007}%
{Shell \MakeLowercase{\textit{et al.}}: Bare Demo of IEEEtran.cls for Computer Society Journals}
\IEEEcompsoctitleabstractindextext{%
\begin{abstract}
Vision--based road detection is an essential functionality for
supporting advanced driver assistance systems (ADAS) such as road
following and vehicle and pedestrian detection. The major
challenges of road detection are dealing with shadows and lighting
variations and the presence of other objects in the scene. Current
road detection algorithms characterize road areas at pixel level
and group pixels accordingly. However, these algorithms fail in
presence of strong shadows and lighting variations. Therefore, we
propose a road detection algorithm based on video alignment. The
key idea of the algorithm is to exploit the similarities occurred
when a vehicle follows the same trajectory more than once. In this
way, road areas are learned in a first ride and then, this road
knowledge is used to infer areas depicting drivable road surfaces
in subsequent rides. Two different experiments are conducted to
validate the proposal on different video sequences taken at
different scenarios and different daytime. The former aims to
perform on--line road detection. The latter aims to perform
off--line road detection and is applied to automatically generate
the ground--truth necessary to validate road detection algorithms.
Qualitative and quantitative evaluations prove that the proposed
algorithm is a valid road detection approach.
\end{abstract}
% Note that keywords are not normally used for peerreview papers.
}
% make the title area
\maketitle

\IEEEdisplaynotcompsoctitleabstractindextext
% \IEEEdisplaynotcompsoctitleabstractindextext has no effect when using
% compsoc under a non-conference mode.

% For peer review papers, you can put extra information on the cover
% page as needed:
% \ifCLASSOPTIONpeerreview
% \begin{center} \bfseries EDICS Category: 3-BBND \end{center}
% \fi
%
% For peerreview papers, this IEEEtran command inserts a page break and
% creates the second title. It will be ignored for other modes.
\IEEEpeerreviewmaketitle

\section{Introduction}
\label{sec:introduction} 
\begin{figure}[htpb!]
\begin{center}
\begin{tabular}{cc}
\includegraphics[width=0.45\columnwidth]{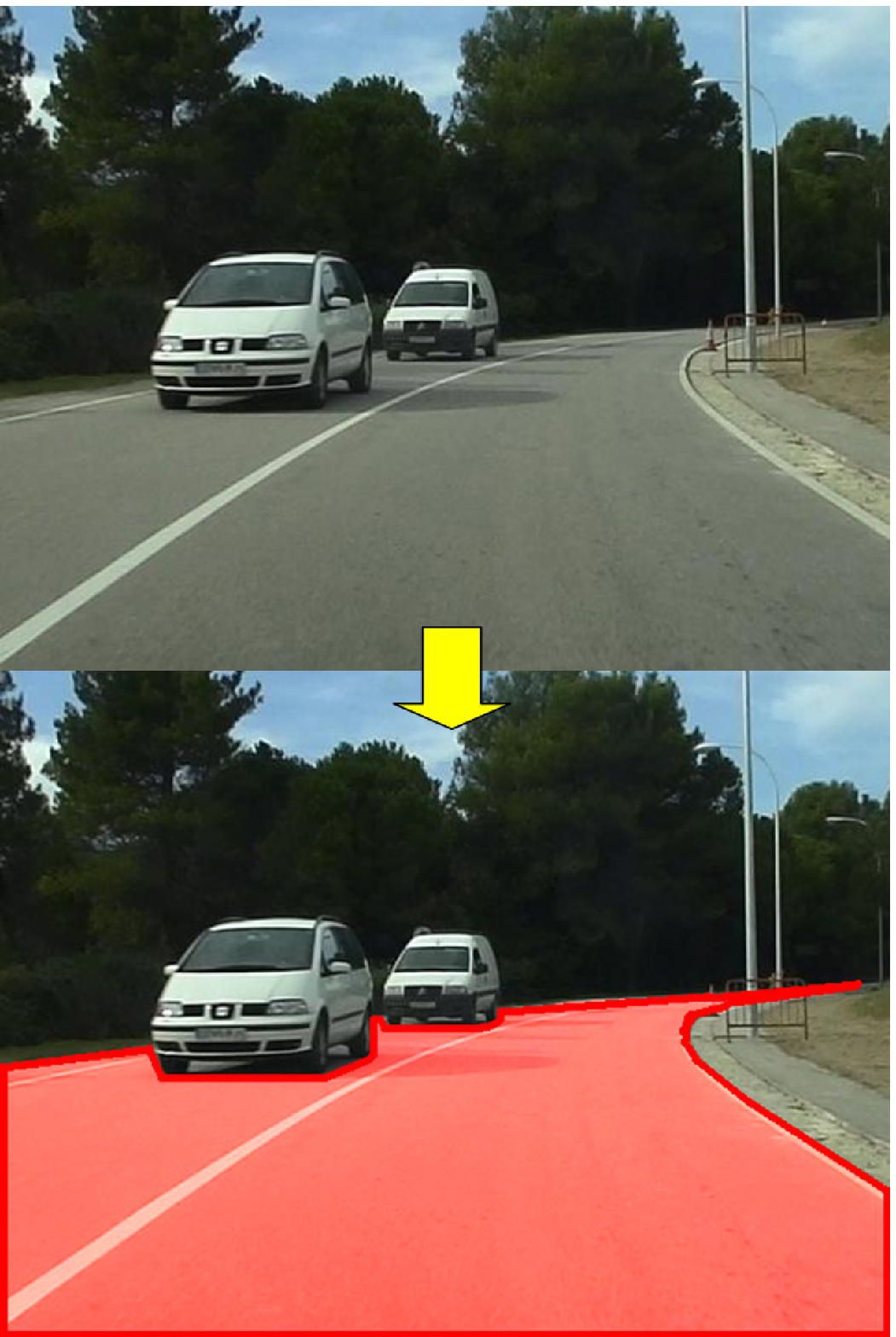} & \includegraphics[width=0.45\columnwidth]{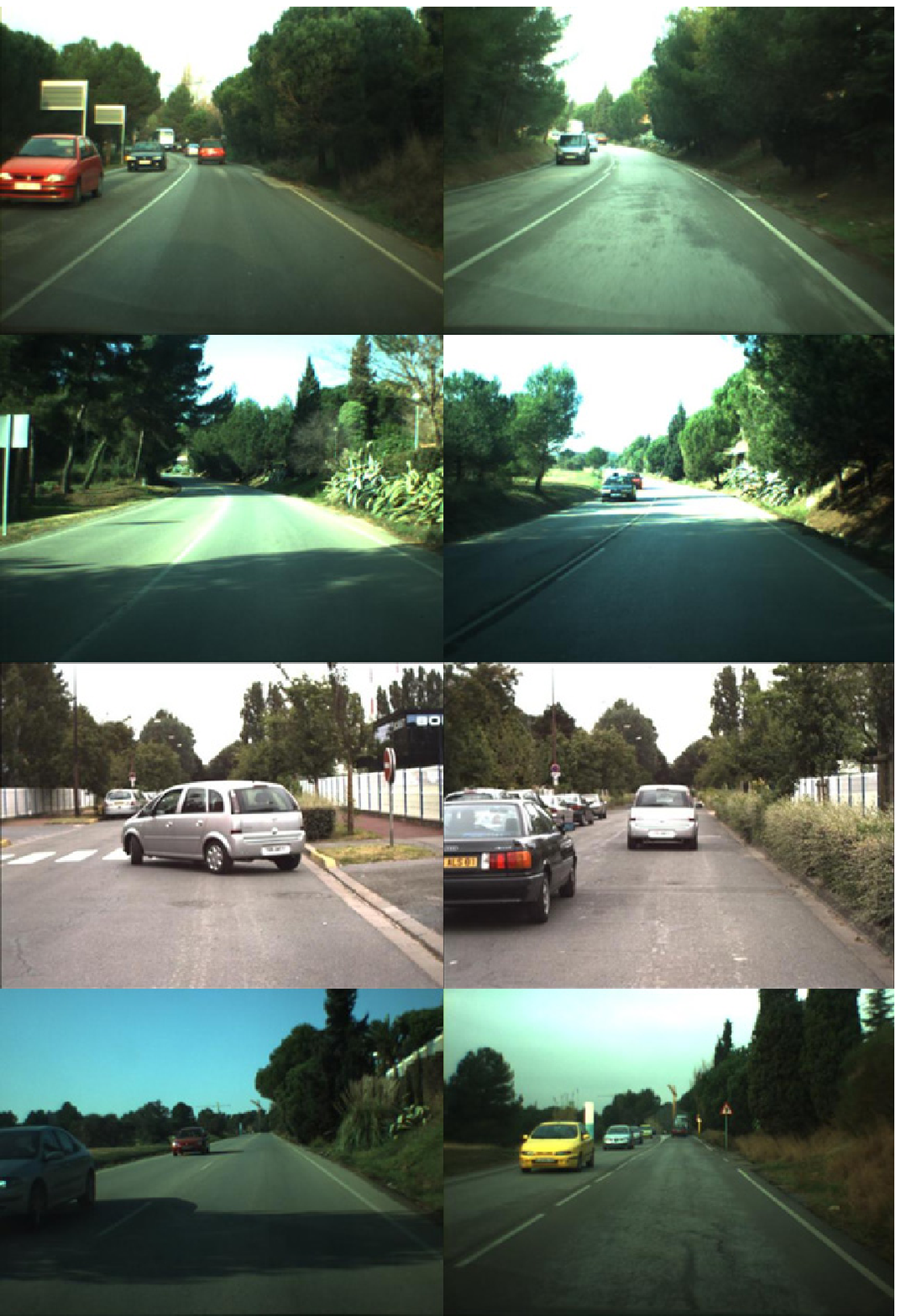}\\
a) & b) \\
\end{tabular}
\end{center}
\caption{(a) Vision--based road detection aims to detect the free
road surface ahead a moving vehicle. (b) The main challenges of
road detection are continuously changing background, the presence
of different objects like vehicles and pedestrian, different road
types (urban, highways, back--road) and varying ambient
illumination and weather conditions.} \label{fig:RD}
\end{figure}

\IEEEPARstart{V}ision--based road
detection aims to detect the free road surface ahead of the
ego--vehicle using an on--board camera~(\subfig{fig:RD}{a}). Road
detection is a key component in autonomous driving to solve
specific tasks such as road following, car collision avoidance and
lane keeping~\cite{Thorpe:1988,Lookingbill:2007}. Moreover, it is
an invaluable background segmentation stage for other
functionalities such as vehicle and pedestrian
detection~\cite{Geronimo:2009}. Road detection is very challenging
since the algorithm must deal with continuously changing
background, the presence of different objects like vehicles and
pedestrian, different road types (urban, highways, back--road) and
varying ambient illumination and weather
conditions~(\subfig{fig:RD}{b}).

\FloatBarrier

Common vision--based road detection algorithms consider road
homogeneity to group pixels according to features extracted at
pixel--level such as texture~\cite{Rasmussen04} and
color~\cite{SoteloITS:2004}. However, algorithms based on
low--level features may fail for severe lighting variations
(strong shadows and highlights) and may depend on structured
roads. The performance of these systems is often improved by
including constraints such as road shape
restrictions~~\cite{SoteloITS:2004} or temporal
coherence~\cite{Tan:2006} at the expense of limiting the
applicability of the algorithm.

\begin{figure}[hpbt!]
\begin{center}
\begin{tabular}{c}
 \includegraphics[width=\columnwidth]{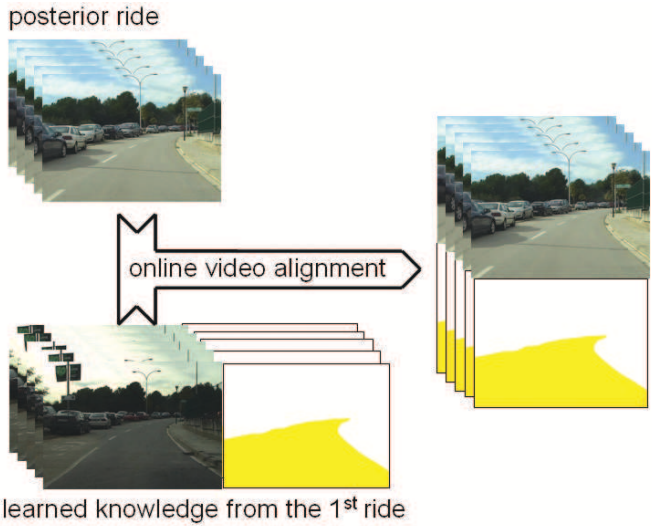}\\
\end{tabular}
\end{center}
\caption{Knowledge learned in a first ride is used to detect road
regions in subsequent rides.} \label{fig:framework}
\end{figure}

In this paper, as a novelty, we propose a road detection approach
based on video alignment. Video alignment algorithms aim to relate
frames and image coordinates between two video
sequences~\cite{Caspi:Irani}. Hence, the key idea of the proposed
algorithm is to exploit similarities occurred when one vehicle
drives through the same route (\ie similar trajectories) more than
once~(\fig{fig:framework}). In this way, road knowledge is learnt
in a first ride and then, video alignment is used to detect the
road in the current image by transferring this knowledge from one
sequence to the current one. The result is a rough segmentation of
the road that is refined to obtain the accuracy required.

The novelty of the paper is twofold: first, we propose an on--line
method to perform video alignment based on image comparisons and a
fixed--lag smoothing approach~\cite{russel03modern}. This method
is specially designed to deal with specific road detection
requirements: independent camera trajectories and independent
vehicle speed variations. Second, a road detection algorithm is
proposed on the basis of on--line video alignment. The algorithm
improve the robustness of video--alignment to shadows by computing
image comparisons in an illuminant--invariant feature space. Then,
this robustness is combined with a refinement step at pixel--level
to achieve the required accuracy.

The rest of this paper is organized as follows: first, in
\sect{sec:SotA} related work is review. Then, in \sect{sec:bg},
the method to perform on--line video alignment approach is
outlined. The algorithm for road detection using on--line video
alignment is described in \sect{sec:RSapproach}. In
\sect{sec:results}, two different experiments are presented to
validate the algorithm. The former aims to perform on--line road
detection. The latter aims to perform off--line road detection and
is applied to automatically generate the ground--truth necessary
to validate road detection algorithms. Finally, in
\sect{sec:conclusions}, conclusions are drawn.

\section{Related Work}
\label{sec:SotA}

\noindent {\bf Vision--based Road Detection}. Road detection
algorithms aim to detect the free road surface ahead of the
ego--vehicle using an on--board camera. Common road detection
algorithms consider road homogeneity to group pixels according to
low--level features (pixel--level) such as
texture~\cite{Lombardi:2005,Rasmussen04} and
color~\cite{Thorpe:1988,SoteloITS:2004,Yinghua:2004}. For
instance, in~\cite{Lombardi:2005}, Lombardi \etal use a
textureless descriptor to characterize road areas. However, the
imaged road texture varies too much with the distance to the
camera due to the perspective effect. In~\cite{Rasmussen04},
Rasmussen \etal use dominant orientations based on Gabor filtering
to detect the vanishing point. However, this approach shows
dependency on strong textures parallel to the road direction in
the form of lane markings for paved roads or tracks left by other
vehicles in rural (unpaved) roads.  In contrast, Kong
\etal~\cite{Kong10a} detect vanishing points using an adaptive
soft--voting scheme based on confidence--weighted Gabor filters.
Color appearance information has been widely accepted as the main
cue for road detection since color imposes less physical
restrictions (regarding the shape of the road), leading to more
versatile systems. The two most popular color spaces, that have
proved to be robust to minor illuminant changes, are $HSV$
\cite{SoteloITS:2004,Rotaru:2008} and normalized $RGB$
\cite{Tan:2006}. However, algorithms based on these color spaces
may fail under wide lighting variations (strong shadows and
highlights among others) and these algorithms depend on highly
structured roads, road homogeneity, simplified road shapes, and
idealized lighting conditions. The performance of these systems is
sometimes improved by including constraints such as temporal
coherence~\cite{Michalke2009,Tan:2006} or road shape
restrictions~\cite{Sotelo:2004}.

\noindent {\bf Video Alignment}. Video alignment algorithms aim to
relate frames and image coordinates between two video
sequences~\cite{carceroni,Caspi:Irani,singh,Tresadern2009891,Vidal,Wolf:Zomet}.
One of these sequences is designated as observed sequence, then
the other is designated as reference. The observed sequence
provides the spatial and temporal reference whereas the reference
one is mapped to match it. Current video alignment focus on
synchronizing sequences simultaneously recorded with fixed or
rigidly attached cameras. These assumptions involve a fixed
spatio--temporal parametric model along the whole sequence. Hence,
the video alignment is posed as a minimization problem over small
amount of parameters comparing some data extracted from the
images. For instance, a common approach consists of computing
image similarities based solely on the gray level
intensity~\cite{Caspi:Irani,Ukrainitz:Irani}. Other
approaches~\cite{singh,Wolf:Zomet} exploit the benefit of temporal
information and track several characteristic points along the
sequences. However, all these approaches are based on rigid camera
attachment and can not deal with the specific requirements of road
detection: $(1)$ independent similar trajectories and $(2)$
independent vehicle speed variations. Another set of works address
the challenge of aligning sequences recorded by independent moving
cameras at different times~\cite{Sand:Teller,diego:2010}. However,
these algorithms require a high computational cost and they can
not be applied to align sequences during acquisition. Therefore,
in the next section, we propose an on--line video alignment based
on a fixed--lag smoothing approach that yields an on--line video
alignment estimation apart from a small fixed delay in processing
the incoming data.

\section{On--line Video Alignment}
\label{sec:bg}

In this section, as a novelty, we propose a video--alignment
approach that is able to estimate spatio--temporal relationship
between two video sequences while one of them is being acquired.
That is, each newly acquired in the observed sequence is mapped
and pixel--wise related to one of the frames in the reference
sequence. The proposed algorithm is based on~\cite{diego:2010} to
deal with road detection requirements. However, there are two
important differences: first, the algorithm requires only a small
number only a small number of frames of the observed sequence to
operate. Second, the algorithm uses a max--product
algorithm~\cite{pearl} instead of using the common Viterbi
algorithm,. The max-product method is a message passing algorithm
that makes direct use of the graph structure in constructing and
passing messages, and is also very simple to implement.

The proposed algorithm consists of two different blocks: on--line
temporal alignment and spatial alignment.

\subsection{On--line temporal alignment}
\label{sec:sincrobg}

On--line temporal alignment, or synchronization, consists of
associating each newly acquired image (in the observed sequence)
to one of the frames in the reference sequence. That is, a single frame is only estimated for each newly acquired image instead of a frame correspondence function. This task is
formulated as a probabilistic labeling problem. A label
$x_t\in\{1,...,n_r\}$ refers the frame number in reference
sequence associated to the $t^{th}$ newly acquired frame. Hence,
the label $x_{t-l}$ is inferred using fixed-lag
smoothing~\cite{russel03modern} on a hidden Markov model that only
considers a small number of frames of the observed sequence as
follows:

\begin{equation}
 x^{\ast}_{t-l} = \argmax_{x_{t-l}\in \Omega_{t}} p(x_{t-l} | \mathbf{y}_{t-L:t}),
 \label{eq:MAPonline}
\end{equation}

\noindent where $x^{\ast}_{t-l}$ is the inferred label at time $t$
for the frame recorded $l$ time units ago. $\Omega_{t}$ is the set
of possible labels, $l\ge0$ is the lag or delay of the system,
$\mathbf{y}_{t-L:t}$ are the  observations from the $(t-L)^{th}$
to $t^{th}$ frame in the observed sequence and $L+1\ge l$ is the
total number of observations available used for inferring the current label $x_{t-l}$.

\begin{figure}[htpb!]
\begin{center}
\includegraphics[width=\columnwidth]{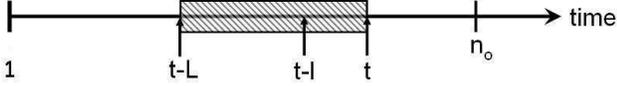}
\end{center}
\caption{Representation of fixed--lag smoothing inference. Label
$x_{t-l}$ is estimated at time $t$ using only $L+1$ frames in the
observed sequence.} \label{fig:fixed}
\end{figure}

The aim of the fixed--lag smoothing is estimating the label
$x_{t-l}$ that must show 'similar content' among the $L+1$ frames
in the observed sequence (\fig{fig:fixed}). Hence, the most likely
adjacent labels of $x_{t-l}$, $\mathbf{x}_{t-L:t}$, and the
observed frames $\mathbf{y}_{t-L:t}$ must also show 'similar
content'. Hence, the $p(x_{t-l} | \mathbf{y}_{t-L:t})$ is
formulated to maximize the frame similarity given the most likely
temporal mapping as follows:

\begin{equation}
p(x_{t-l} | \mathbf{y}_{t-L:t}) = \max_{\mathbf{x}_{t-L:t}\backslash x_{t-l}}{p(\mathbf{x}_{t-L:t} | \mathbf{y}_{t-L:t})},
 \label{eq:fixed}
\end{equation}

\noindent where $\mathbf{x}_{t-L:t}=[x_{t-L},\ldots,x_t]$ is the
temporal mapping between the reference sequence and ${t-L:t}$
frames in the observed sequence, $\mathbf{x}_{t-L:t}\backslash x_{t-l}$ considers all variables except to $x_{t-l}$, and $p(\mathbf{x}_{t-L:t} |
\mathbf{y}_{t-L:t})$ measures the temporal correspondence between
the observed frames and a reference sequence. Furthermore, the
posterior probability density in \eq{eq:fixed} is decoupled as
follows:

\begin{eqnarray}
p(\mathbf{x}_{t-L:t} | \mathbf{y}_{t-L:t})& \propto & p(\mathbf{y}_{t-L:t} | \mathbf{x}_{t-L:t}) p(\mathbf{x}_{t-L:t}),
\label{eq:fixed_decouple}
\end{eqnarray}

\noindent where $p(\mathbf{y}_{t-L:t} | \mathbf{x}_{t-L:t})$ and
$p(\mathbf{x}_{t-L:t})$ are the observation likelihood and the
\emph{prior}, respectively. This prior $p(\mathbf{x}_{t-L:t})$
favors only labellings that satisfies some assumptions (\eg the
vehicles can stop independently), whereas the observation
likelihood $p(\mathbf{y}_{t-L:t} | \mathbf{x}_{t-L:t})$ measures
the similarity between a pair of videos given a temporal mapping
$\mathbf{x}_{t-L:t}$. Finally, the max--product algorithm is used
to infer the label $x^{\ast}_{t-l}$ as follows:

\begin{equation}
 x^{\ast}_{t-l} = \argmax_{x_{t-l}\in \Omega_{t}} \max_{\mathbf{x}_{t-L:t}\backslash x_{t-l}}{p(\mathbf{y}_{t-L:t} | \mathbf{x}_{t-L:t}) p(\mathbf{x}_{t-L:t})}.
 \label{eq:maxproduct}
\end{equation}

For simplicity, the prior $p(\mathbf{x}_{t-L:t})$ and the
observation likelihood $p(\mathbf{y}_{t-L:t} |
\mathbf{x}_{t-L:t})$ are factorized as follows:

\begin{eqnarray}
p(\mathbf{x}_{t-L:t}) & = & P(x_{t-L})\prod_{k=t-L}^{t-1}p(x_{k+1} |x_k)
\end{eqnarray}

\noindent and

\begin{eqnarray}
p(\mathbf{y}_{t-L:t} | \mathbf{x}_{t-L:t}) & = & \prod_{k=t-L}^{t}p(y_k|x_k),
\end{eqnarray}

\noindent under the assumption that the transition and the
observation probabilities are conditionally independent given the
previous and current label values, respectively. $P(x_{t-L})$ gives the same probability to all labels $\Omega_t$ to avoid the propagation of possible errors in temporal assignments.

The intended meaning of $p(x_{k+1} |x_k)$ is that vehicles do not
go backward, that is, they move always forward or at most stop for
some time. Therefore, labels $x_t$ must increase monotonically as
follows:
\begin{eqnarray}
p(x_{k+1}\mid x_{k}) &=& \left\{
\begin{array}{ll}
\beta & \textrm{if} \ x_{k+1} \ge x_k \\
0 & \textrm{otherwise}
\end{array} \right.,
\label{eq:constraint}
\end{eqnarray}

\noindent where $\beta$ is a constant that gives the same
importance to all label configurations $\mathbf{x}_{t-L:t}$
satisfying the constraint in \eq{eq:constraint}. That \emph{prior} $p(x_{k+1}\mid x_{k})$ does not restrict the vehicle speed, but they can vary independently.

The intended meaning of $p(y_k|x_k)$ is to measure a frame
similarity given a pair of frames and is defined as follows:

\begin{equation}
%p(y_k|x_k) = \max_{\substack{-\Delta_x<i<\Delta_x \\ -\Delta_y<j<\Delta_y}}{\Phi(f(y_k,y_{x_k}^{i,j}); \mu_y,\sigma_y^2)},
p(y_k|x_k) = \Phi(f(y_k,y_{x_k}); \mu_y,\sigma_y^2),
\label{eq:app_lik}
\end{equation}

\noindent where $y_k$ and $y_{x_k}$ are the image descriptors of
the $k^{th}$ and the $x_k^{th}$ frames in the observed and
reference sequence respectively, $f(y_k,y_{x_k}^{i,j})$ is a similarity measure between both descriptors, and $\Phi(v;\mu_y,\sigma_y^2)$
denotes the evaluation of the Gaussian pdf
$\mathcal{N}(\mu,\sigma_y^2)$ at $v$, being $\mu_y$ and $\sigma_y$ the mean and variance of the similarity measure $f(\cdot,\cdot)$.

%to minimize the influence of lighting variations and a

\subsection{Spatial alignment}
\label{sec:spatialbg}

Spatial alignment consists of estimating a geometric
transformation that relates the image coordinates of a pair of
corresponding frames. For any such pair, the cameras are assumed
to be at the same position but their $3D$ orientation (pose) may
be different because of trajectory differences, and acceleration,
braking and road surface irregularities  affecting the yaw and
pitch angles, respectively. Hence, the geometric transform
existing between two corresponding frames is an special class of
homography, the conjugate rotation $H = K R K^{-1}$, being $K =
\mbox{diag}(f,f,1)$ and $f$ the focal length of the camera in
pixels. It is important to bear in mind that, despite this
notation does not express it for sake of simplicity, this
transformation is not constant along the whole sequence. The
transformation changes for every pair of corresponding frames thus
making difficult the synchronization task. The rotation matrix $R$
expresses the relative orientation of the cameras for one pair of
corresponding frames, and it is parametrized by the Euler angles
${\mathbf{\Omega}}= (\Omega_x, \Omega_y, \Omega_z)$ (pitch, yaw
and roll respectively). Furthermore, the transformation modeled by
$H$ is approximated using a quadratic motion model as
follows~\cite{Irani}:

\begin{eqnarray}
{\mathcal{W}({\mathbf x};{\mathbf{\Omega}})} = \left[
\begin{array}{rrr}
-\frac{xy}{f}\; & \;f+\frac{x^2}{f} & \;-y \\
-f-\frac{y^2}{f}\; & \frac{xy}{f} \; & \;x  \\
\end{array}
\right] \left[
\begin{array}{c}
\Omega_x \\
\Omega_y \\
\Omega_z \\
\end{array}\right].
\label{eq:angles}
\end{eqnarray}

In this way ${\mathbf{\Omega}}$ is estimated minimizing the sum of
squared differences by means of the additive forward
implementation of the Lukas--Kanade algorithm~\cite{Simon:Baker}
as follows:

\begin{eqnarray}
{\mathbf{\Omega^*}} = \argmin{\mathbf{\Omega}} \big( \sum_{\mathbf
x} \big[ S_{x_t}^r({\mathbf x}+ \mathcal{W}({\mathbf x};{
\mathbf{\Omega}}))-S_{t}^o({\mathbf x}) \big]^2 \big),
\label{eq:Err}
\end{eqnarray}

\noindent where $S_{x_t}^r$ is the image warped onto the image
coordinates of $S_t^o$. $\mathbf{\Omega}$ is iteratively estimated
in a coarse--to--fine manner. For a detailed description we refer
the reader to \cite{Simon:Baker}.

\section{Road Detection based on Video Alignment}
\label{sec:RSapproach}

In this section, a novel road detection algorithm based on
on--line video alignment is proposed. The algorithm consists of
two stages: on--line video alignment and refinement. Thus, the
proposed algorithm combines the robustness of video alignment to
provide road segmentations despite lighting conditions and the
accuracy of a pixel--level refinement process. The first stage
relates frame-- and pixel--wise two video sequences and transfers
road knowledge from the first sequence to the current ride.
Further, the algorithm improves robustness against lighting
variations and shadows by using a shadowless feature space. The
second stage is a based on dynamic background subtraction to
remove objects in the observed sequence. This refinement consists
of analyzing road regions based on the image dissimilarities
between both rides. Thus, the process assumes the stored sequence
is recorded with the absence or low--density traffic.

\subsection{On--line video alignment for road detection}
\label{sec:on_line_va}

\begin{figure}[h!]
\begin{center}
\includegraphics[width=\columnwidth]{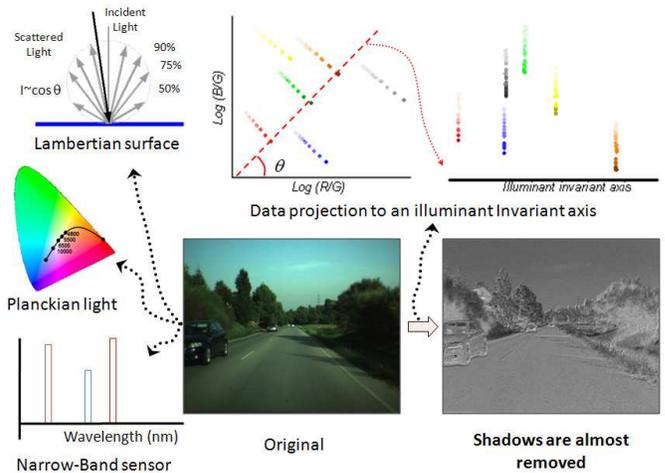}
\end{center}
\caption{An illuminant--invariant image is obtained under the
assumptions of Planckian light, Lambertian surface and narrow-band
sensors. This image is almost shadow free.}
\label{fig:Assumptions}
\end{figure}

The first stage of the algorithm consists of applying the on--line
video alignment method~(\sect{sec:bg}). Moreover, robustness
against lighting variations and shadows is improved by using an
illuminant--invariant feature space~\cite{Finlayson:2006} to
perform image comparisons~(\eq{eq:app_lik}). This
illuminant--invariant space minimizes the influence of lighting
variations under the assumption of \emph{Lambertian} surfaces
imaged by a three fairly narrow-band sensor under approximately
Planckian light sources~\cite{Finlayson:2006}. As shown in
\fig{fig:Assumptions}, the characterization process (\ie
converting an RGB image onto the shadow--less feature space)
consists of projecting the $\{log(R/G)$, $log(B/G)\}$ pixel values
of the image onto the direction orthogonal to the lighting change
lines, \emph{invariant--direction} $\theta$. This direction is
device dependent and can be estimated off-line using the
calibration procedure of~\cite{Finlayson:2006}.

\begin{figure}[h!]
\begin{center}
\includegraphics[width=\columnwidth]{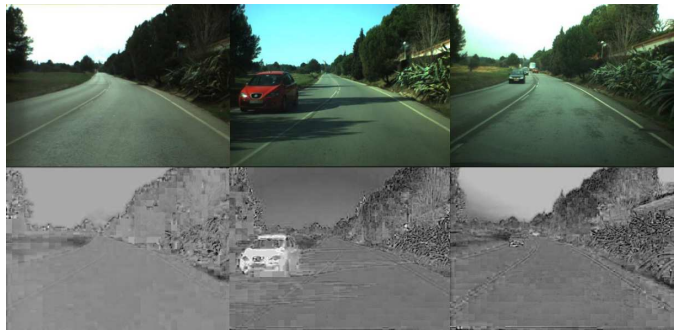}
\end{center}
\caption{Illuminant--invariant examples of images acquired
approximately at the same position under different lighting
conditions.} \label{fig:InvariantExample}
\end{figure}

In practice, image descriptors $y_{\ast}$ are computed as follows.
First, the image converted onto the shadow--less feature space
shown in \fig{fig:InvariantExample} is smoothed using a Gaussian
kernel with $\sigma$ and downsampled along each axis at
$1/16^{th}$ of the original resolution. Then, partial derivatives
are computed setting the gradient magnitude at each pixel equal to
zero if it is less than $5\%$ of the maximum. The reason to employ
a low threshold of the gradient magnitude instead of the intensity
value itself is to reinforce to the lighting invariance
conditions. Finally, all the partial derivatives are stacked into
a column vector that is normalized to unit norm. Then, the
similarity measure $f(\cdot,\cdot)$ is defined as the maximum of
the inner product between descriptors among different horizontal
and vertical translations of the smoothed downsampled input image,
which are set up to $2$ pixels. Thus, $\Phi(v;\mu_y,\sigma_y^2)$
is related to the closest coincidence angle between two descriptor
vectors, whose $\mu_y$ and $\sigma_y$ are set empirically to $1$
and $0.5$. That maximum makes the similarity measure in
\eq{eq:app_lik} invariant, to some extent, to slight rotations and
translations between the reference and observed frames. These
dissimilarities are unavoidable because, of course, the vehicle
will not follow exactly the same trajectory in the two rides.

\begin{figure}[h!]
\begin{center}
\begin{tabular}{cc}
\includegraphics[width=0.45\columnwidth]{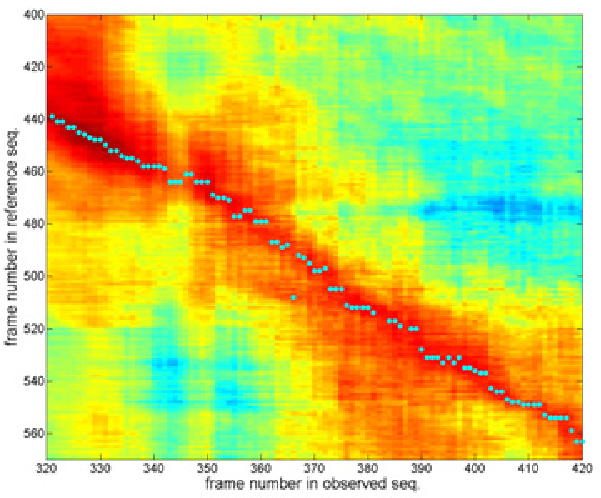} & \includegraphics[width=0.45\columnwidth]{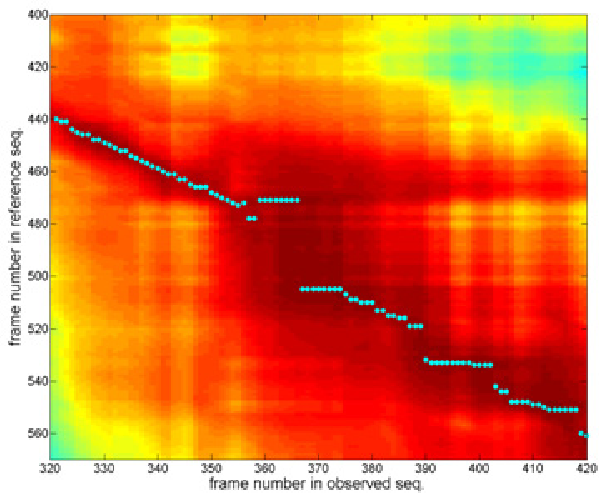}\\
(a) & (b) \\
\end{tabular}
\end{center}
\caption{Using an illuminant--invariant feature space improves the
performance of the on--line synchronization algorithm. Images show
on--line synchronization (filled circles) on a background
inversely proportional to frame similarity in~\eq{eq:app_lik}. a)
Synchronization results using the illuminant--invariant feature
space. b) Synchronization results using gray level images.}
\label{fig:Invariant_comparison}
\end{figure}

Figure~\ref{fig:Invariant_comparison} shows the synchronization
benefits of using the illuminant--invariant feature space.
Further, quantitative evaluation results in a lower average
synchronization error when the illuminant--invariant feature space
is used ($1.05 \pm 0.8791$ against $1.75\pm 0.87$). From this
results we can conclude that using the illuminant--invariant
representation improves the accuracy of the algorithm to
discriminate corresponding frames.

\subsection{Road refinement}
\label{sec:refinement}

The result of the video alignment stage is a pair of corresponding
frames $(t,x_t)$ and its relative geometric transformation
$\mathbf{\Omega}_t$ that relates them pixel--wise. Hence, road
regions $\mathcal{M}_{t}$ delimiting the road surface in the
observed sequence at time $t$ are obtained by warping the road
segmentation of the corresponding frame in the reference sequence
$\mathcal{M}_{x_t}$ using $\mathbf{\Omega}_t$ as follows:

\begin{equation}
\mathcal{M}_{t} = \mathcal{M}_{x_t}(\mathbf{x}+\mathcal{W}({\mathbf x};{
\mathbf{\Omega}_t})).
\end{equation}

However, the transferred road regions $\mathcal{M}_{t}$ is a rough
approximation of the free--road surface due to the observed
sequence may show different objects (\eg vehicles, pedestrian).
Therefore, the refinement algorithm removes regions that contain
those objects in the observed frames and are within the
transferred road regions. This assumption is because the detected
objects are claimed to the observed sequence due to the fact that
the first ride is recorded with the absence or low--density
traffic. Therefore, a dynamic background subtraction is proposed
to detect objects spotting differences between a corresponding
frame pair.

\begin{figure}[h!]
 \begin{center}
 \begin{tabular}{cc}
 \includegraphics[width=0.4\columnwidth]{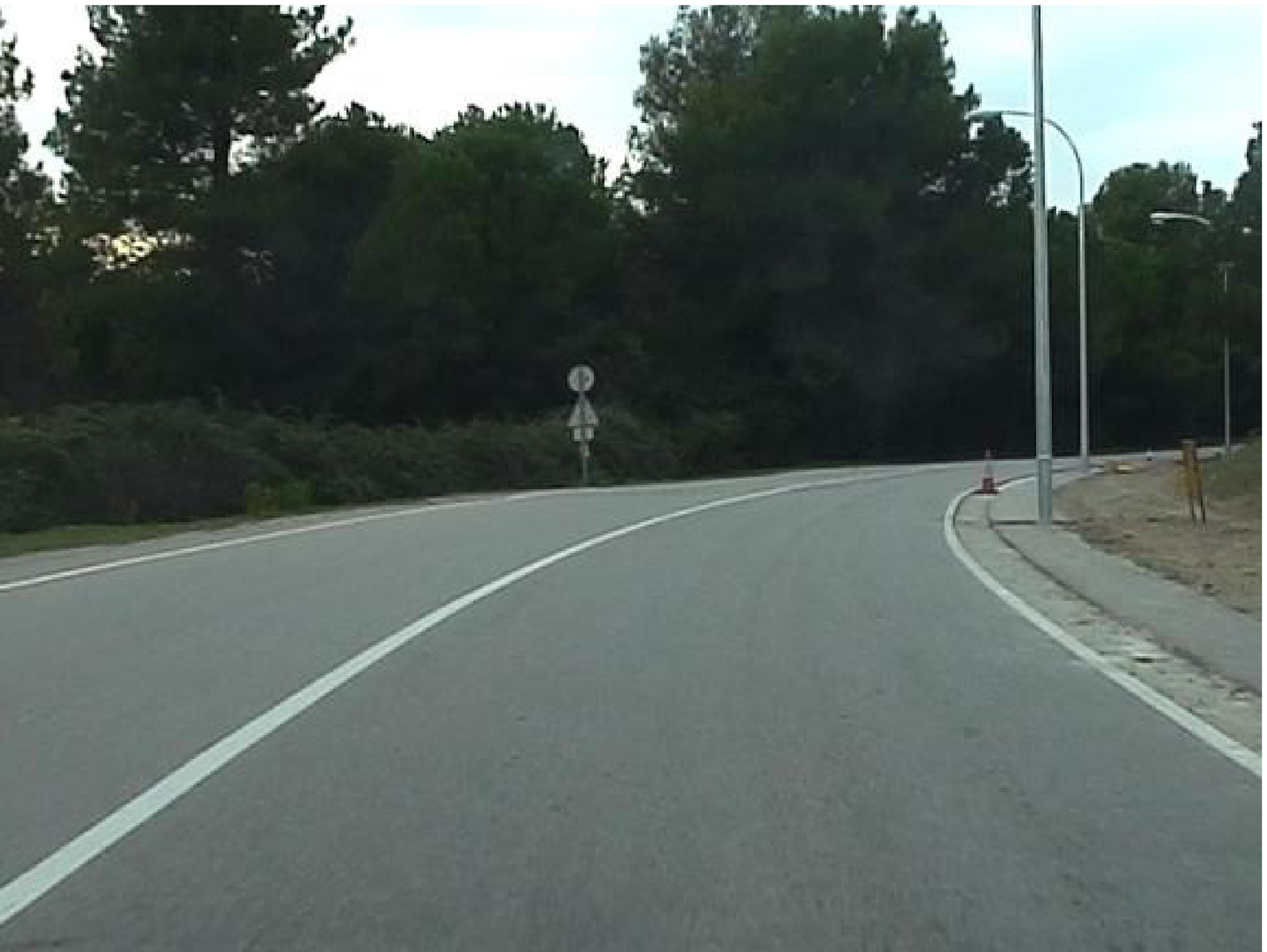}&
 \includegraphics[width=0.4\columnwidth]{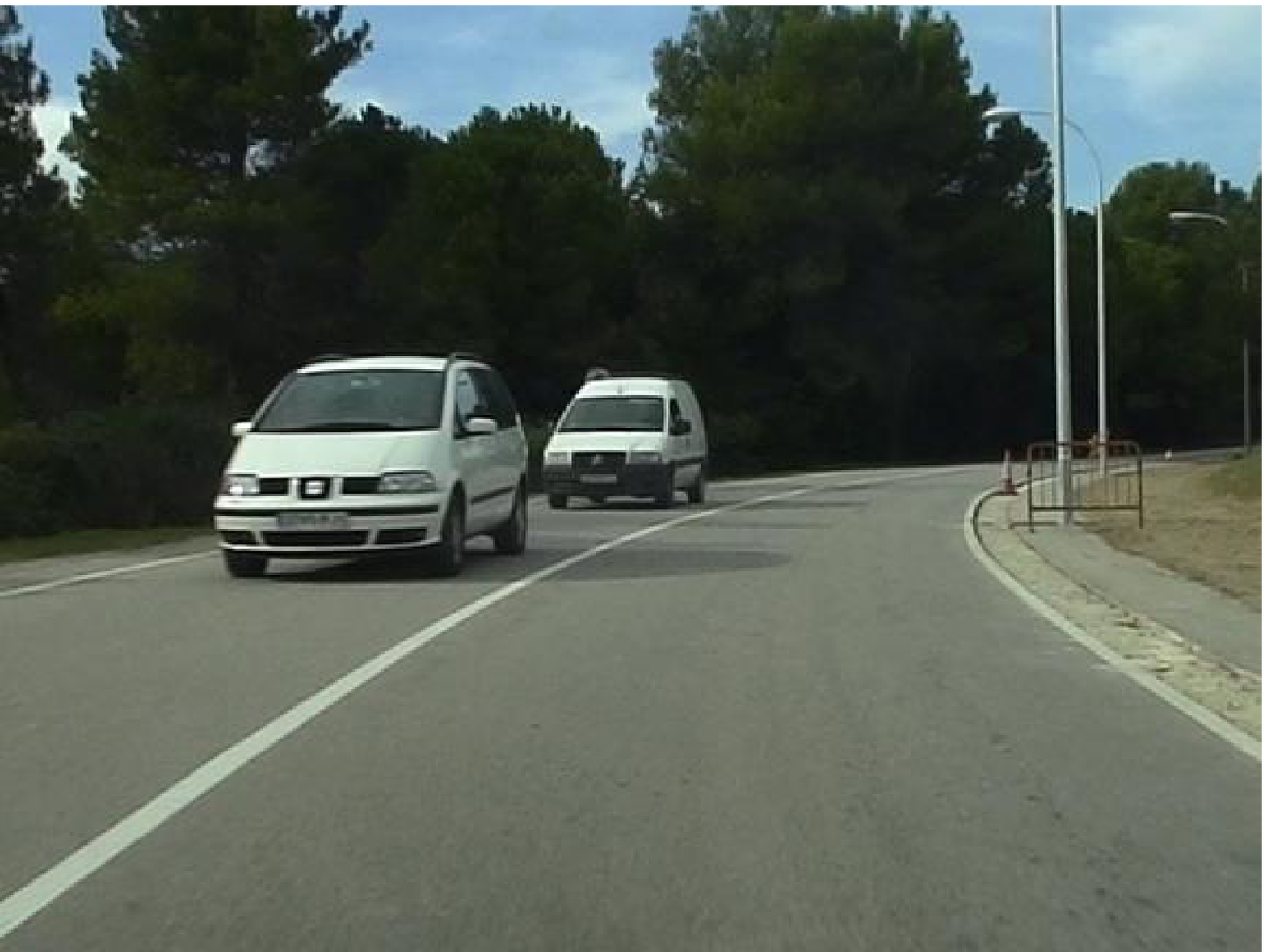}\\
 (a)&(b)\\
 \includegraphics[width=0.4\columnwidth]{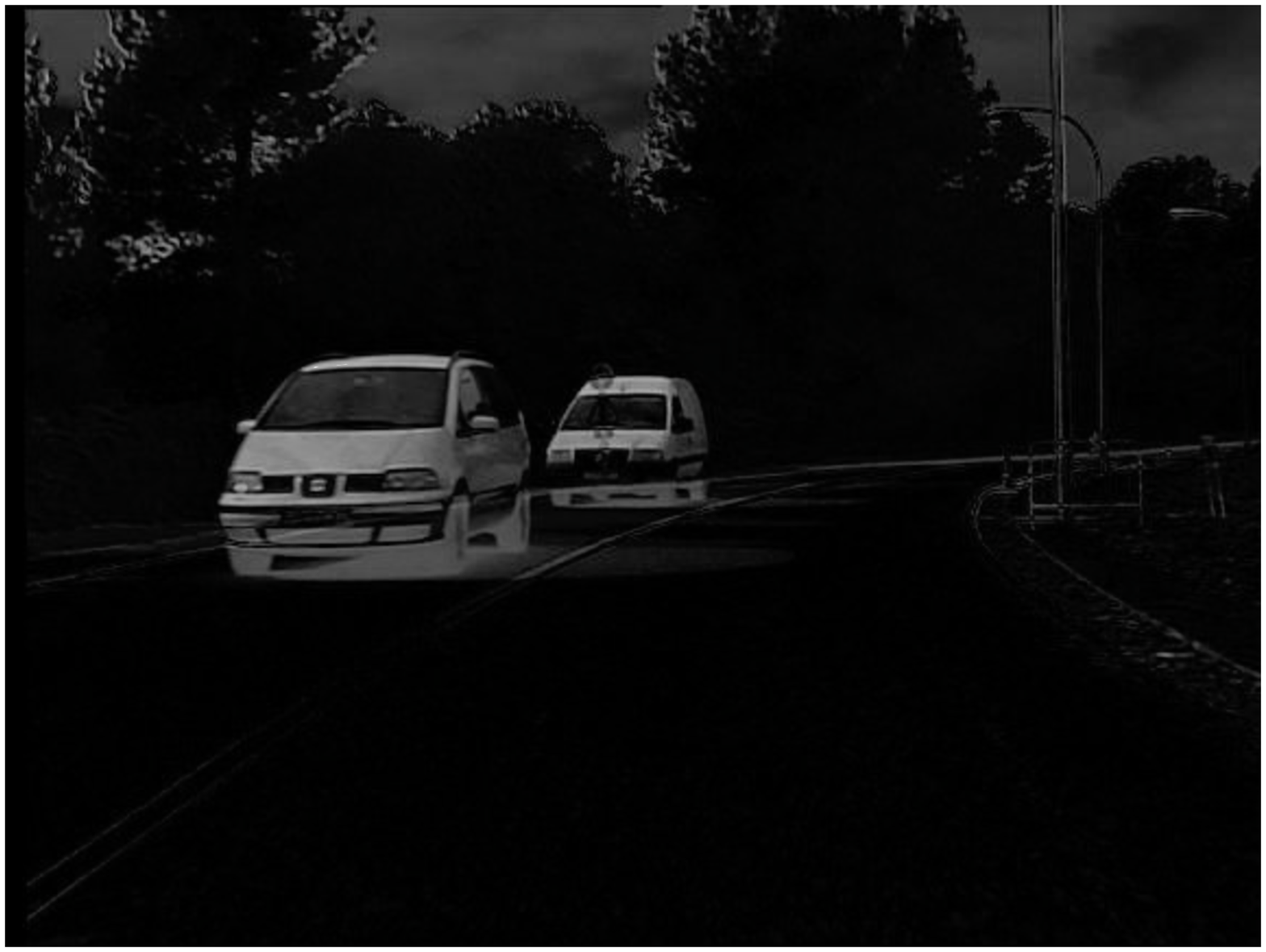}&
 \includegraphics[width=0.4\columnwidth]{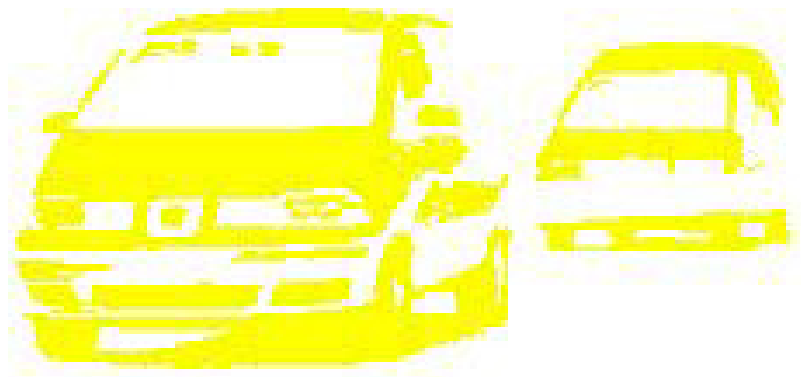}\\
 (c)&(d)\\
 \includegraphics[width=0.4\columnwidth]{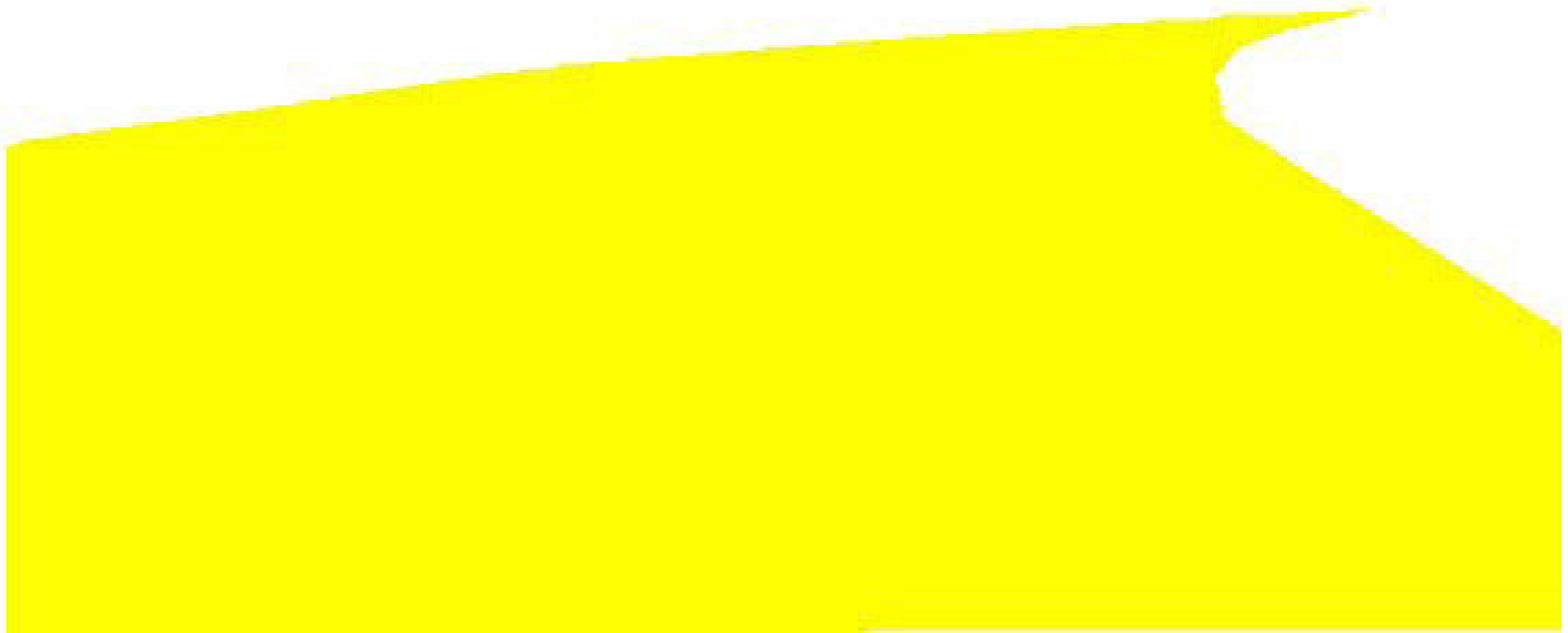}&
 \includegraphics[width=0.4\columnwidth]{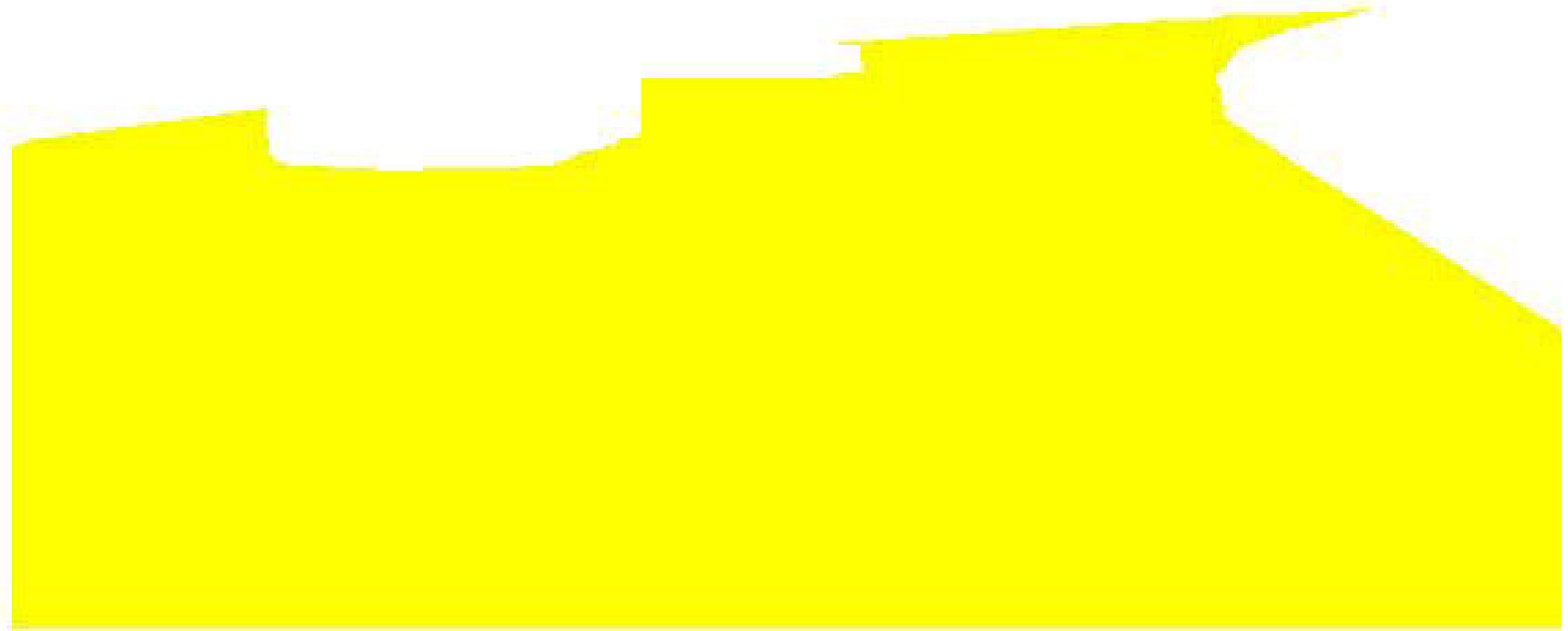}\\
 (e)&(f)\\
 \end{tabular}
 \end{center}
 \caption{Refinement shows the corresponding frame from the
 database (a) is aligned with the input frame (b). The difference
 between them (c) is used to detect foreground objects (d). The
 known road surface of the corresponding frame is transferred to
 the input image (e), and finally, the foreground vehicles are
 removed (f).}\label{fig:refinement}
 \end{figure}

In particular, the dynamic background subtraction is computed as
follows (\fig{fig:refinement}). First, the corresponding frame in
the reference sequence $S_{x_t}^{r}$ is warped to the image
coordinates of the observed frame, $S_{x_t}^{r,w}$. Then, the
intensity of a pair of corresponding frames are subtracted
pixel--wise as $S_{x_t}^{r,w}-S_{t}^{o}$, being $S_t^{o}$ the
observed frame at time $t$. Hence, the subtraction allows to spot
differences of potential interest that are considered as objects
present in the observed sequence. Hence, the detected regions are
considered forward objects. Then, the absolute value of the pixel
subtraction, $\mid S_{x_t}^{r,w}-S_{t}^{o} \mid$, is binarized
using automatic thresholding techniques~\cite{otsu_ieeetsmc_1979},
and the possible holes in the binary regions are filled using
mathematical morphology. \fig{fig:refinement} illustrates an
example of refinement procedure to remove regions that contains
vehicles and are within the transferred road.

\section{Experiments}
\label{sec:results}

In this section, two different experiments are conducted to
validate the proposed algorithm on different sequences acquired with a forward--facing camera attached at the windscreen. The goal in the first experiment
is detecting free--road areas ahead of a vehicle. The second
experiment consist of applying the algorithm to automatically
generate ground--truth to evaluate the performance of road
detection algorithms.

\subsection{Datasets}
Experiments are conducted on three different scenarios: 'street',
'back--road' and 'campus'. The first two scenarios consist of three video sequence pairs; whereas 'campus' scenarios only one video sequence pair. The following pairs, 'Street--1' and 'street--2' provided by~\cite{Kong10b}, and 'back--road--1' and 'back--road--2', are three video sequences recorded following the same route, one reference and two observed sequences, to demonstrate the robustness of inferring free--drivable areas under different lighting conditions. The observed sequence in 'back--road--1', 'back--road--3', 'street--2', 'street--3' and reference in 'campus' are recorded at noon in a sunny day
under the presence of shadows on the road surface; whereas the observed sequence in 'street--1' is recorded in a sunny day at morning under the presence of shining road surface. The rest of sequences do not contain shadows because the reference sequences except in 'campus' are recorded in a cloudy day whereas the observed sequence in 'campus' was acquired during the sunset and the observed sequence in 'back--road--3' is acquired in a cloudy day with a wet road surface. Furthermore, both sequences in 'street--1' and 'street--3' scenario are free of vehicles in contrast with 'back--road--1' and 'back--road--1' that contain vehicles in both sequences. The rest of sequence pairs deal with the presence of vehicles in the observes sequence. The number of frames in observed sequence differs from the reference sequences due to differences in the trajectory and speed of the vehicle. \tab{tab:scenario} summarizes the main characteristics of each scenario and sequences to demonstrate the
variability of the experiments. All sequences are recorded at the
same frame rate. The road regions of the reference and observed
sequence in all scenarios are manually delineated.

\begin{table*}[htbp]
\caption{Description of the main characteristics of the scenarios
and sequences.} \label{tab:scenario}
%\begin{minipage}{\columnwidth}
\begin{center}\begin{tabular}{|c|c||c|c|c|c|c|}\hline
\multirow{2}{*}{\small{Scenario}} &\multirow{2}{*}{\small{Sequence}}& \small{Recording} &\multirow{2}{*}{\small{weather}}   &       \multirow{2}{*}{\small{Shadows}}     &\multirow{2}{*}{\small{Vehicles}} &                     \multirow{2}{*}{\small{length}}      \\
 & & \small{time}      &    && & \\ \hline \hline
\multirow{2}{*}{\small{Back--Road--1}} & \small{Observed}  &  \small{noon}     & \small{sunny}  & \small{yes} & \small{yes} & \small{$714$} \\ \cline{2-7}
          & \small{Reference} & \small{morning} & \small{cloudy} & \small{no}  & \small{yes} & \small{$948$} \\ \hline\hline

\multirow{2}{*}{\small{Back--Road--2}} & \small{Observed}  &  \small{noon}     & \small{cloudy}  & \small{wet} & \small{yes} & \small{$402$} \\ \cline{2-7}
            & \small{Reference} & \small{morning} & \small{cloudy} & \small{no}  & \small{yes} & \small{$948$} \\ \hline\hline

\multirow{2}{*}{\small{Street--1~\cite{Kong10a}}} & \small{Observed}  &  \small{noon}     & \small{sunny}  & \small{yes} & \small{yes} & \small{$210$} \\ \cline{2-7}
                                                 & \small{Reference} & \small{noon} & \small{cloudy} & \small{no}  & \small{no} & \small{$239$} \\ \hline\hline

\multirow{2}{*}{\small{Street--2~\cite{Kong10a}}} & \small{Observed}  &  \small{morning}     & \small{sunny}  & \small{shining} & \small{no} & \small{$260$} \\ \cline{2-7}
                                                     & \small{Reference} & \small{noon} & \small{cloudy} & \small{no}  & \small{no} & \small{$239$} \\ \hline\hline

\multirow{2}{*}{\small{Street--3}} & \small{Observed}  &  \small{noon}     & \small{sunny}  & \small{yes} & \small{no} & \small{$520$} \\ \cline{2-7}
                                                 & \small{Reference} & \small{afternoon} & \small{cloudy} & \small{no}  & \small{no} & \small{$627$} \\ \hline\hline
\multirow{2}{*}{\small{Back--Road--3}} & \small{Observed}  &  \small{noon}      & \small{sunny}  & \small{yes} & \small{yes} & \small{$1318$} \\ \cline{2-7}
                                                    & \small{Reference} &  \small{afternoon} & \small{cloudy} & \small{no}  & \small{no}  & \small{$1459$} \\ \hline \hline
\multirow{2}{*}{\small{Campus}} & \small{Observed}  &  \small{sunset} & \small{sunny} & \small{no}  & \small{yes} & \small{$600$} \\ \cline{2-7}
                                                 & \small{Reference} &  \small{noon}   & \small{sunny} & \small{yes} & \small{no}  & \small{$816$}  \\ \hline
\end{tabular}\end{center}
%\end{minipage}
\end{table*}

\subsection{Performance Evaluation}
Quantitative evaluations are provided using pixel--based measures
defined in a contingency table (\tab{tab:contingency}). The
entries of this table are defined as follows: TP is the number of
correctly labelled road pixels, TN is the number of non-road
pixels detected, FP is the number of non-road pixels classified as
road pixels and FN is the number of road pixels erroneously marked
as non-road. Further, using the entries in the contingency table,
the following error measures are computed: accuracy, sensitivity,
specificity and quality, see~\tab{tab:measures}. Each of these
measures provides different insight of the results. Accuracy
provides information about the fraction of classifications that
are correct. Specificity measures the proportion of true negatives
(\ie background pixels) which are correctly identified.
Sensitivity, or recall, is  the ratio of detecting true positives
(\ie road pixels). Quality is related to the completeness of the
extracted data as well as its correctness. All these measures
range from $0$ to $1$, from worst to perfect.

\begin{table}[htpb]
\caption{The contingency table. Algorithms are evaluated based on
the number of pixels correctly and incorrectly classified. See
text for entries definition.} \label{tab:contingency}
\begin{center}
\begin{tabular}{|p{0.08cm}|p{0.5cm}|p{0.5cm}|c|}
\hline \multicolumn{2}{|c|}{\scriptsize{Contingency}} & \multicolumn{2}{c|}{\scriptsize{Ground--truth}} \\
\cline{3-4} \multicolumn{2}{|c|}{\scriptsize{Table}} &\scriptsize{Non--Road} & \scriptsize{Road} \\
\hline \multirow{2}{*}{{\begin{sideways}\scriptsize{Result}\end{sideways}}}& \scriptsize{Non--Road} & \footnotesize{TN}& \footnotesize{FN} \\
\cline{2-4}   & \scriptsize{Road} & \footnotesize{FP} & \footnotesize{TP}\\
\hline
\end{tabular}
\end{center}
\end{table}

\begin{table}[htpb]
\caption{Pixel--wise measures used to evaluate the performance of
the different detection algorithms. These measures are defined
using the entries of the contingency table
(\tab{tab:contingency}). } \label{tab:measures}
  \centering
\renewcommand{\multirowsetup}{\centering}
\vspace{2mm}
\begin{center}
\begin{tabular}{|m{1.9cm}|l|}
\hline \centering \scriptsize{Measure} & \multicolumn{1}{c|}{\scriptsize{Definition}} \\
\hline  \scriptsize{Quality} & $\hat{g} =
\frac{TP}{TP+FP+FN}$\\
\hline  \scriptsize{Accuracy} &
\footnotesize{$ACC=\frac{TP+TN}{TP+FP+FN+TN}$}
\\\hline \scriptsize{Sensitivity} & \footnotesize{$TPR=\frac{TP}{TP+FN}$} \\
\hline \scriptsize{Specificity} & \footnotesize{$SPC=\frac{TN}{FP+TN}$}\\
\hline
\end{tabular}
\end{center}
\end{table}
\subsection{On--line Road Detection Results}
\begin{figure*}[htbp!]
\begin{center}
\includegraphics[width=\textwidth,height = 23cm]{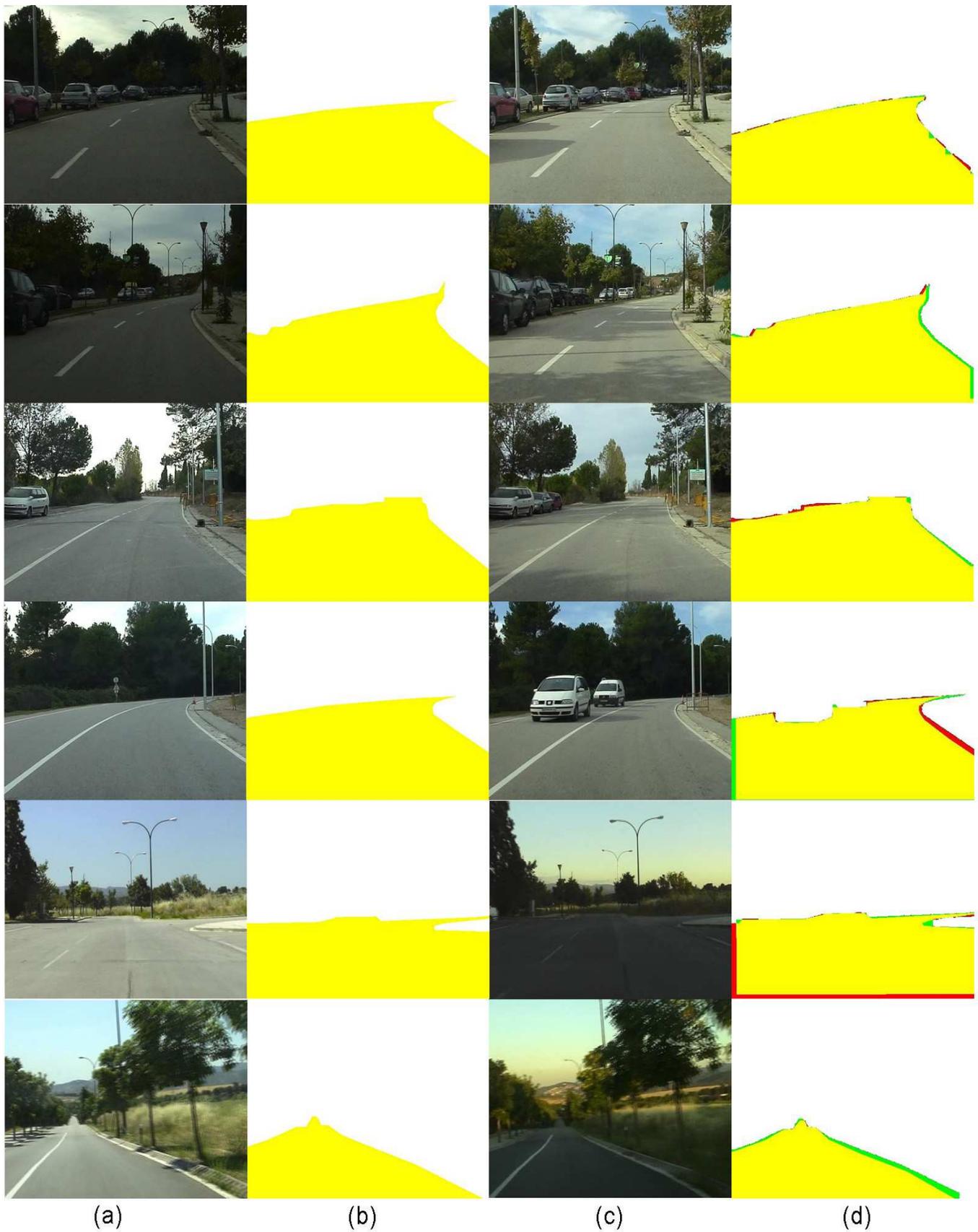}  %\\
\end{center}
\caption{Example results of the proposed road detection algorithm
for different scenarios. The frame from the reference sequence (a)
is aligned with the input frame (c). Learned road regions (b)
combined with the refinement stage are used to generate the final
result (d). The color code in the image is as follow: true
positives are in yellow; true negatives are in white; false
positives are in read and, false negatives are in green, with
respect to a road/non--road classification. More results, in video
format, can be viewed at
\emph{http://www.cvc.uab.es/$\sim$fdiego/RoadSegmentation/}.}\label{fig:results}
\end{figure*}

\begin{figure*}[htbp!]
\begin{center}
\includegraphics[width=\textwidth,height = 23cm]{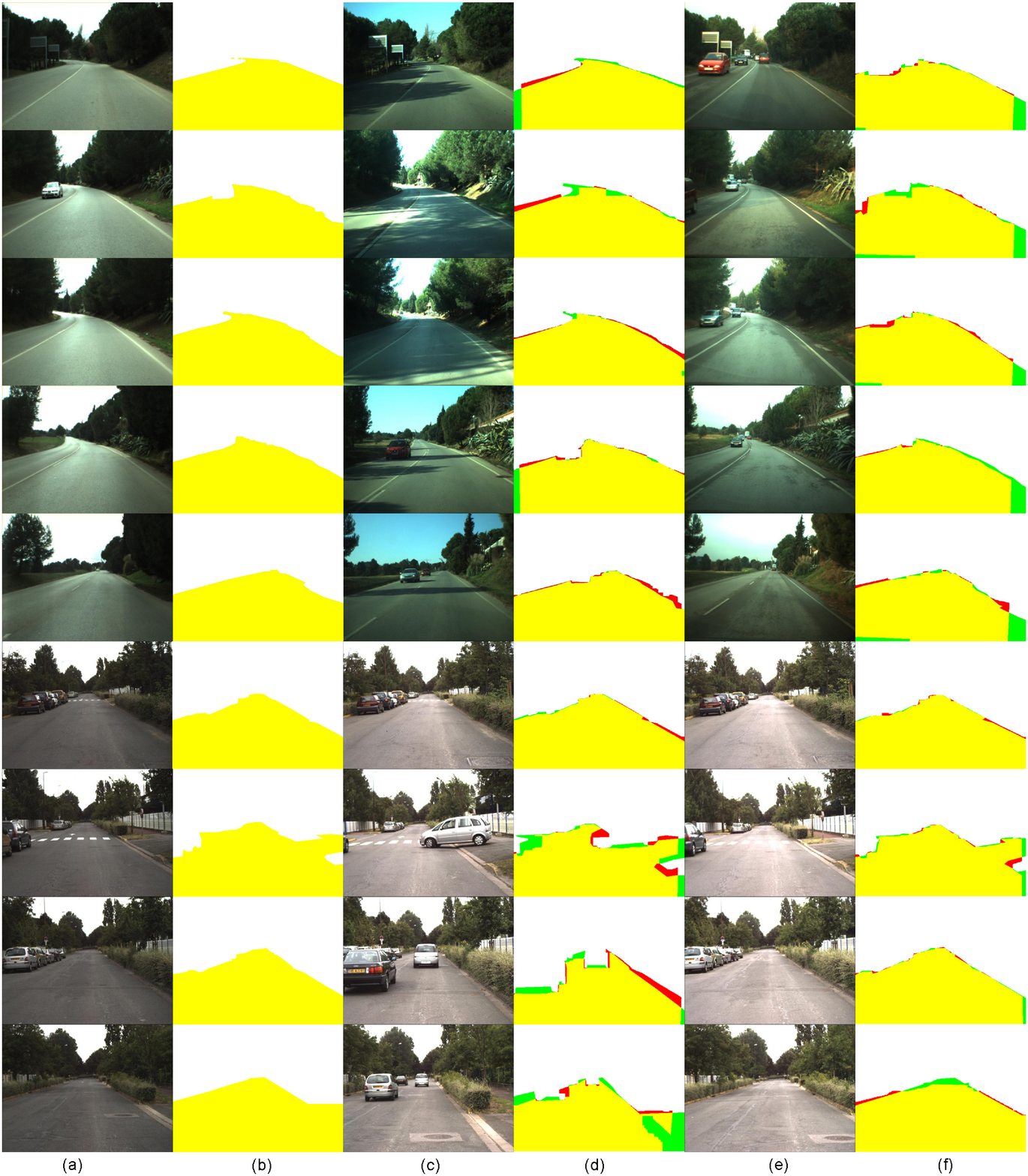}  %\\
\end{center}
\caption{Example results of the proposed road detection algorithm
for two different scenarios driven, at least, 3 times. The same frame from the reference sequence (a)
is aligned with the input frames (c) and (e) under different lighting conditions. Learned road regions (b)
combined with the refinement stage are used to generate the final
results (d) and (f). The color code in the image is as follow: true
positives are in yellow; true negatives are in white; false
positives are in read and, false negatives are in green, with
respect to a road/non--road classification.}\label{fig:results2}
\end{figure*}

In the first experiment, the significance of including the
refinement stage is evaluated comparing the performance before and
after the refinement stage. A summary of quantitative evaluations
is listed \tab{tab:Quantitative1} and some example results are
shown in \fig{fig:results} and \fig{fig:results2}. As shown, road
areas are properly transferred from the reference sequence to
observed sequence. Specifically, \fig{fig:results2} shows the
robustness of transferring the same \emph{road prior} to different
observed sequences under different lighting conditions. That is,
the proposed algorithm deals with the presence of shadows and a
wet road surface (the first five rows in
\subfig{fig:results2}{c-d} and \subfig{fig:results2}{e-f},
respectively), and shining road surface (the last four rows in
\subfig{fig:results2}{c-f}). Errors (shown in red and green in
\fig{fig:results}) and \fig{fig:results2}) are mainly located at
the road boundary mainly due to the ambiguity of manually
delimiting the road boundaries. Furthermore, the refinement step
handles correctly the presence of vehicles cropping properly the
transferred road region. That step increases the performance in
all four measures as shown in \tab{tab:Quantitative1}. In
addition, \subfig{fig:results2}{c-f} shows the benefits (larger
discriminative power) of including the refinement step when other
vehicles are present in the scene. This is mainly due to the fact
that the refinement stage removes on--coming, in--coming or parked
vehicles. From these results we can conclude that the proposed
algorithm is able to recover road areas despite different lighting
conditions, \eg shadows, wet and shining surface, and the presence
of vehicles with different size, colors and shapes in the scene.
Finally, as shown in the second row of \subfig{fig:results2}{a-b},
the proposed algorithm also handles the presence of vehicles
(low--dense traffic) in the reference sequence reducing the
accuracy since the with vehicles are not transferred. This is also
reinforced quantitatively in \tab{tab:Quantitative1} where the
performance in 'back--road--1' and 'back--road--2' is comparable
to the performance of other pairs of sequences.

\begin{table*}[htbp]
\caption{Average performance of the proposed road detection
algorithm over all the corresponding frames.
}\label{tab:Quantitative1}
%\begin{minipage}{\columnwidth}
\begin{center}\begin{tabular}{|c|c||c|c|c|c|}\hline
\small{Scenario}  &  \small{Refinement}              &
\small{$\hat{g}$} & \small{$SPC$}& \small{$TPR$} & \small{$ACC$}
\\ \hline \hline \multirow{2}{*}{\small{Back--Road--1}}  & without
& $0.9637\pm 0.03792$ & $0.9724\pm0.0285$ &
$\mathbf{0.9528\pm0.05489}$ & $0.9640\pm 0.0400$ \\ \cline{2-6}
                                       & within  & $\mathbf{0.9680\pm0.0.0304}$ & $\mathbf{0.9819\pm0.0111}$ & $0.9504\pm0.0624$ & $\mathbf{0.9760\pm0.0163}$ \\ \hline \hline

\multirow{2}{*}{\small{Back--Road--2}}  & without &
$0.9425\pm0.02364$ & $0.9688\pm0.0238$ &
$\mathbf{0.9105\pm0.0312}$ & $0.9597\pm0.0308$ \\ \cline{2-6}
                                                          & within  & $\mathbf{0.9467\pm0.0193}$ & $\mathbf{0.9834\pm0.0138}$ & $0.9018\pm0.0374$ & $\mathbf{0.97793\pm0.0187}$ \\ \hline \hline

\multirow{2}{*}{\small{Street--1~\cite{Kong10a}}}  & without &
$0.9415\pm0.0380$ & $0.9348\pm0.0507$ & $\mathbf{0.9556\pm0.0438}$
& $0.9057\pm0.0795$ \\ \cline{2-6}
                                                          & within  & $\mathbf{0.9495\pm0.0351}$ & $\mathbf{0.96001\pm0.0343}$ & $0.9356\pm0.0617$ & $\mathbf{0.9367\pm0.0601}$ \\ \hline \hline

\multirow{2}{*}{\small{Street--2~\cite{Kong10a}}}  & without &
$0.9828\pm0.0194$ & $0.9897\pm0.0125$ & $0.9744\pm0.0386$ &
$0.9878\pm0.0132$ \\ \cline{2-6}
                                                          & within  & $\mathbf{0.9846\pm0.0190}$ & $\mathbf{0.9903\pm0.0094}$ & $\mathbf{0.9778\pm0.0388}$ & $\mathbf{0.9885\pm0.0102}$ \\ \hline \hline

\multirow{2}{*}{\small{Street--3}}  & without & $0.9691\pm0.0117$
& $0.9914\pm0.0044$ & $0.9807\pm0.0099$ & $0.9869\pm0.0051$ \\
\cline{2-6}
                                                          & within  & $\mathbf{0.9817\pm0.0092}$ & $\mathbf{0.9936\pm0.0039}$ & $\mathbf{0.9904\pm0.0092}$ & $\mathbf{0.9923\pm0.0039}$ \\ \hline \hline

\multirow{2}{*}{\small{Back--Road--3}} & without &
$0.9559\pm0.0546$ & $0.9869\pm0.0258$ & $0.9728\pm0.0236$ &
$0.9802\pm0.0262$ \\\cline{2-6}
                                                                        & within & $\mathbf{0.9626\pm0.0543}$ & $\mathbf{0.9909\pm0.0244}$ & $\mathbf{0.9746\pm0.0260}$ & $\mathbf{0.9832\pm0.0262}$ \\ \hline \hline

\multirow{2}{*}{\small{Campus}} & without & $0.9127\pm0.0890$ &
$0.9897\pm0.0075$ & $0.9403\pm0.0842$ & $0.9766\pm0.0219$ \\
\cline{2-6} & within & $\mathbf{0.9262\pm0.0794}$ &
$\mathbf{0.9909\pm0.0078}$ & $\mathbf{0.9501\pm0.0713}$ &
$\mathbf{0.9794\pm0.0236}$ \\\hline
\end{tabular}\end{center}
%\end{minipage}
\end{table*}
An inherent limitation of the method is the delay before obtaining
the results. This delay is set exactly to $200ms$ (\ie $5$ frames
at 25fps). However, this is a minor limitation if a high
frame--rate camera is provided.

\subsection{Off-line road detection: Automatic Ground--truthing}

Ground-truth data is a must for the quantitative assessment and
comparison of detection/segmentation algorithms. In the context of
road detection, the manual annotation of the road regions on
sequences hundreds or thousands frames long is very time consuming
and prone to error because of the human operator's attention drop
off. The required effort is even higher for works that claim to be
robust to different lighting conditions like~\cite{Alvarez:2008},
since for one same track there are several sequences that must be
manually annotated. Thus, automatic generation of ground-truth for
evaluating road detection algorithms is a problem of interest in
itself.

\begin{figure}[htpb!]
\begin{center}
\begin{tabular}{cccc}
\includegraphics[width=0.45\columnwidth]{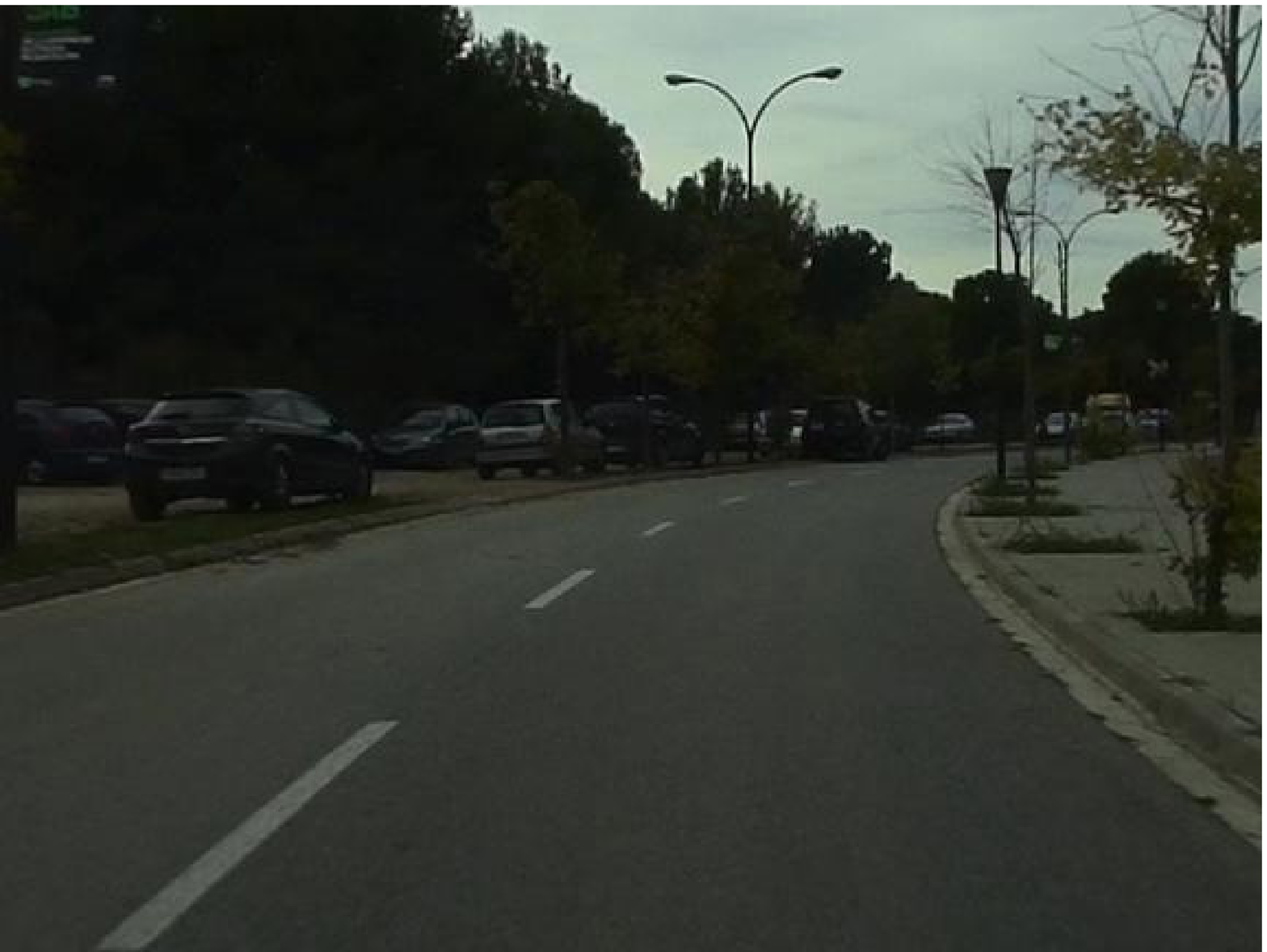}&
\includegraphics[width=0.45\columnwidth]{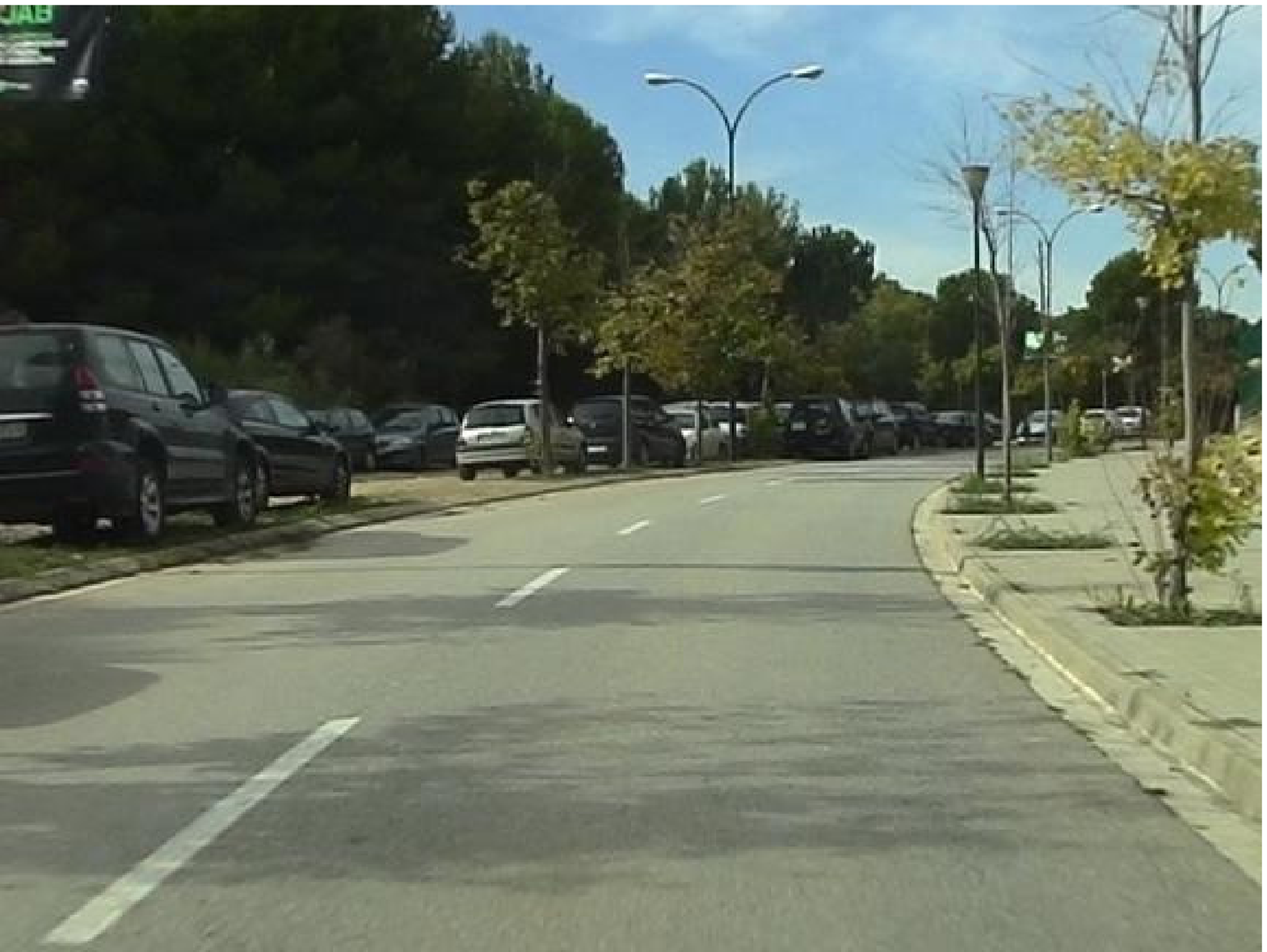}\\(a)&(b)\\
\includegraphics[width=0.45\columnwidth]{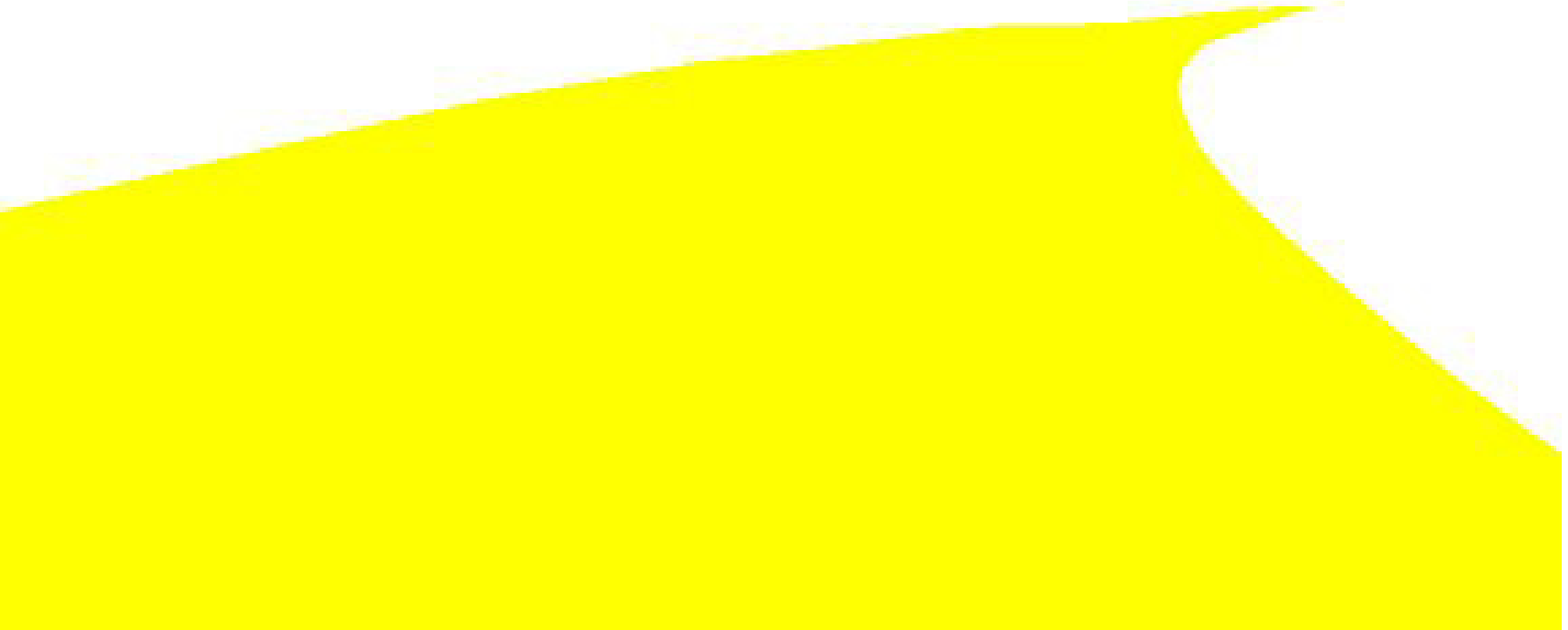}&
\includegraphics[width=0.45\columnwidth]{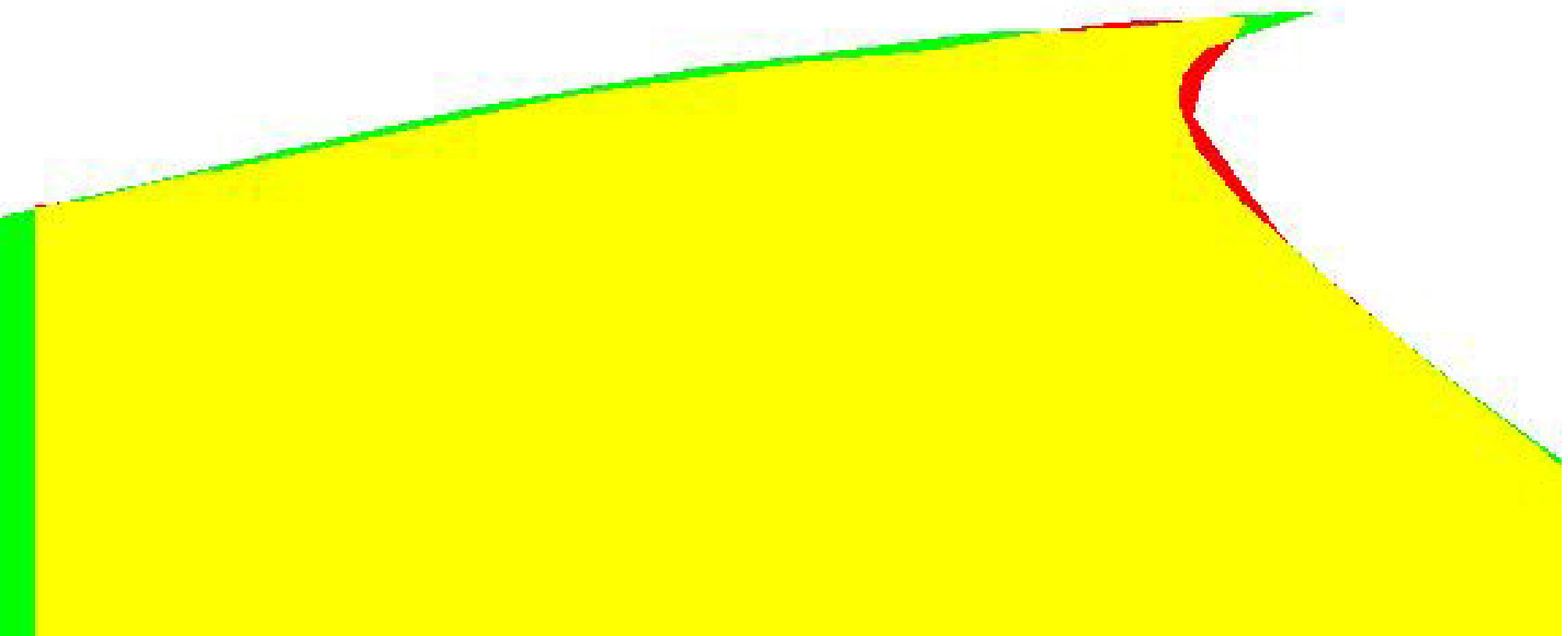}\\
(c)&(d)\\
\end{tabular}
\end{center}
\caption{Example results of the proposed automatic
ground--truthing algorithm. The frame from the learned sequence
(a) is aligned with the new acquired frame (b). The reference
ground--truth (c) is used to generate the output ground--truth
(d). Yellow color refers to true positive pixels. White means true
negative pixels, red false positives and green false
negatives.}\label{fig:results1}
\end{figure}

The proposed algorithm for automatic ground--truthing transfers
the manual annotation in one sequence to another when they are
completely recorded. Example results are shown in
\fig{fig:results1}. More results in video format can be viewed at
\emph{http://www.cvc.uab.es/$\sim$fdiego/RoadSegmentation/}. These
results suggest that the learned ground--truth in the reference
sequence is correctly transferred to the observed one. As shown in
\subfig{fig:results1}{d}, errors  are mainly located at the road
boundaries. However, these errors are due to the inherent boundary
ambiguity when the images are manually segmented by an human
operator.

\begin{table*}[htbp]
\caption{Performance of the ground--truthing algorithm conducted
on 'parking' scenario. }\label{tab:Quantitative}
%\begin{minipage}{\columnwidth}
\begin{center}\begin{tabular}{|c|c|c|c|c|}\cline{2-5}
\multicolumn{1}{c|}{} &\small{$\hat{g}$} &
\small{$SPC$}&\small{$TPR$}&\small{$ACC$}\\\hline
\small{Reference Seq. as reference}& $0.9784\pm0.0103$ & $0.9928\pm0.0057$ & $0.9882\pm0.0095$ & $0.9909\pm0.0044$ \\ \hline
\small{Observed Seq. as reference} & $0.9748\pm0.0535$ & $0.9914\pm0.0068$ & $0.9867\pm0.0538$ & $0.9894\pm0.0218$ \\ \hline
\end{tabular}\end{center}
%\end{minipage}
\end{table*}
Quantitative evaluations are summarized in \tab{tab:Quantitative}.
Two different evaluations are conducted on 'parking' scenario to
demonstrate the capacity of generating accurate ground--truth
using any traffic--free sequence as a reference sequence. The
former keeps the same nomenclature of sequences whereas the latter
interchanges the reference sequence as an observed sequence and
vice versa. The averaged performance over all the corresponding
frames is shown in \tab{tab:Quantitative}. Labelling an image
takes $30$ seconds in average time so using the algorithm on
'parking' sequences saves $4.3$ and $3.7$ hours, respectively.
Small differences are due to the different number of frames in
each video sequence. The highest performance is achieved when the
largest video sequence is used as reference. The main reason is
that the algorithm does not interpolate the information between
frames. Thus, the large amount information available as reference,
the highest accuracy in the registration process. However, this is
a minor drawback since the reference sequence could be recorded
driving at a lower speed or at a higher frame--rate.

An inherent limitation of the method is the presence of moving
vehicles in the reference sequence. However, it is a minor
drawback because the road regions occluded by vehicles can be
interpolated according to the available road boundaries.  Further,
the algorithm can be used in semi--supervised mode. That is, the
ground--truth is automatically generated and shown to the operator
for validation.

\section{Conclusion}
\label{sec:conclusions} In this paper on--line video alignment has
been introduced for road detection. The key idea of the algorithm
is to exploit similarities occurred when a vehicle follows the
same route more than once. Hence, road knowledge is learnt in a
first ride and then, this knowledge is used to infer road areas in
subsequent rides. Furthermore, a dynamic background subtraction is
proposed to handle correctly the presence of vehicles cropping
properly the inferred road region. Thus, the algorithm combines
the robustness against local lighting variations of video
alignment with the accuracy at pixel--level provided by the
refinement step. Experiments are conducted on different image
sequences taken at different day time on real--world driving
scenarios. From qualitative and quantitative results, we can
conclude that the proposed algorithm is suitable for detecting the
road despite varying lighting conditions (\ie, shadows and
different daytime) and the presence of other vehicles in the
scene.
\bibliographystyle{IEEEbib}
\bibliography{references_dissertation}

% that's all folks
\end{document}